\documentclass[11pt]{article}
\usepackage[margin=1in]{geometry}
\usepackage{amsmath,amssymb,amsthm,booktabs,hyperref}
\hypersetup{hidelinks,
 pdftitle={Candidate-Set Sensitivity of Proximal Pessimistic Policy Selection},
 pdfauthor={Soheil Sayah Varg, Mahdi Shirinbayan, Mahdi Ostadmohammadi, Amirhomayoon Sharifizade}}

\newtheorem{proposition}{Proposition}
\newcommand{\Pn}{P_{\mathcal N}}

\newcommand{\OracleBCAtLarge}{0.0363}
\newcommand{\LearnedPessAtLarge}{0.0363}

\newcommand{\CandidateGainAtLarge}{0.14511}
\newcommand{\CandidateGainHalfInterval}{0.03484}
\newcommand{\MaximumRestartGap}{0.0003871}
\title{Candidate-Set Sensitivity of Proximal Pessimistic Policy Selection}
\author{\begin{tabular}{@{}c@{\hspace{2.5em}}c@{}}Soheil Sayah Varg & Mahdi Shirinbayan \\ Mahdi Ostadmohammadi & Amirhomayoon Sharifizade\end{tabular}}
\date{November 11, 2026}
\begin{document}
\maketitle
\begin{abstract}
We study a tabular implementation of proximal policy evaluation followed by
pessimistic selection in confounded partially observable Markov decision
processes. In one synthetic environment, a calibrated, class-free ellipsoid
selector becomes worse as the sample size increases, while a plug-in selector
using the same fitted bridges improves. This finding depends strongly on the
candidate set: adding a behaviour clone learned from logged data reduces
the largest-sample mean regret from $0.1814$ to $0.0363$ on a common
augmented-set reference over 20 simulation seeds. The matched plug-in has mean
regret $0.00002$ on that set. A blockwise null-space calculation explains a possible source
of persistent penalties, but does not imply an impossibility of policy ranking.
Across five new environments with fixed calibration and disjoint trajectory
seeds, adding the behaviour-cloned candidate reduces the largest-sample mean
regret of pessimistic selection from $0.2607$ to $0.0396$. The corresponding
plug-in mean is $0.0327$, so its near-zero regret in the
original environment does not generalize.
Reversing the reward action columns while retaining the logger reverses the
direction of the candidate intervention in four of five environments.
A one-decision illustration using public randomized stroke-trial outcomes
finds no resolved value benefit under artificially induced training selection.
We give the correct pairwise comparison and identify implementation and
reporting conditions needed to interpret these experiments. The evidence is a
case study of a heuristic implementation outside identification assumptions,
not a refutation of proximal reinforcement learning guarantees.
\end{abstract}

\section{Question and scope}
When logged actions depend on a latent state, observation-only policy
evaluation can be confounded. Hong, Qi and Xu~\cite{hong} identify reward and
transition effects using bridge functions and propose pessimistic optimization
over confidence regions intersected with a bounded bridge class. We implement
the tabular bridge estimator and ask a narrower empirical question: can a
pessimism layer worsen selection on fixed fits, and how much does the candidate
set change that finding?

The experiment deliberately uses an inadequate negative control. It therefore
lies outside the completeness assumptions supporting the original guarantees.
Moreover, our principal selection runs omit the bounded bridge class and use
coordinate descent for the coupled minimization. We call the resulting scores
\emph{heuristic pessimistic scores}, rather than certified lower confidence
bounds. Conclusions apply to this implementation and environment.

Pessimism's dependence on data coverage is established in offline
RL~\cite{jin}; it is not a new general mechanism discovered here. Our useful
observation is the size of a candidate-set intervention on otherwise matched
fits. Nonidentification and null-space orthogonality are also established
inverse-problem facts~\cite{babii}. The following calculations provide context,
not a claim of a novel general impossibility theorem.

\paragraph{Relation to prior work.}
Rank or completeness conditions support proxy identification~\cite{miao,tchetgen}
and proximal off-policy evaluation~\cite{bennett}. Regularization under
nonidentification~\cite{babii} provides the geometric context.

In confounded POMDPs, Venkatesh and Malikopoulos~\cite{venkatesh} compare
cross-fitting with a single split in a first-step observation-transition
recovery experiment. Wang et al.~\cite{wang} fit neural reward and dynamics
bridges and use model rollouts for policy optimization. Kuang et
al.~\cite{kuang} measure value-estimation error for prespecified policies with
a distinct tabular estimator.

Related offline-RL work studies distribution shift and coverage~\cite{levine,jin},
comparisons with behaviour cloning~\cite{kumar}, and ranking under
uncertainty~\cite{yang}. Kallus and Zhou~\cite{kalluszhou} optimize
confounding-robust improvement relative to a baseline. Our experiment holds
the fitted bridges and per-seed regret reference fixed and measures how
adding a behaviour-cloned candidate from the logged data changes heuristic
pessimistic selection. A score-error calculation characterizes when candidate
addition can increase regret. We test sensitivity in new environments and
under reward reversal.

\section{Block geometry and policy comparisons}
Freeze every bridge block except one. Its empirical regularized loss has
quadratic difference $(b-\hat b)^\top H(b-\hat b)$, where
$H=A+\lambda I$, $A\succeq0$, $\lambda>0$, and radius $\xi\ge0$.
We write $\lVert g\rVert_{H^{-1}}=\sqrt{g^\top H^{-1}g}$.
The corresponding ellipsoid is
\begin{equation}
 C_\xi=\{b:(b-\hat b)^\top H(b-\hat b)\le\xi\}.
\end{equation}
For the blockwise linear functional $V_g(b)=g^\top b$, its exact minimum is
$g^\top\hat b-W_g$, with $W_g=\sqrt{\xi}\|g\|_{H^{-1}}$.
Let $\mathcal N=\ker A$, with orthogonal projector $\Pn$.

\begin{proposition}[An ellipsoid width requirement]
If $\Pn\hat b=0$ and $b_\star\in C_\xi$, then
\begin{equation}
 W_g\ge\|\Pn b_\star\|\,\|\Pn g\|.
 \label{eq:floor}
\end{equation}
\end{proposition}
\begin{proof}
Define the absolute null-space norms
$\beta=\lVert\Pn b_\star\rVert_2$ and
$\beta_g=\lVert\Pn g\rVert_2$; neither is a normalized fraction.
Because $A\succeq0$ and $\Pn\hat b=0$, coverage implies
\[
 \xi\ge(b_\star-\hat b)^\top H(b_\star-\hat b)
 \ge\lambda\beta^2.
\]
On $\mathcal N$, $H$ acts as $\lambda I$; on its orthogonal complement,
$H^{-1}$ is positive definite. Hence
$g^\top H^{-1}g\ge\beta_g^2/\lambda$. Therefore
\begin{equation*}
 \begin{aligned}
 W_g&=\sqrt{\xi\,g^\top H^{-1}g}\\
    &\ge\sqrt{(\lambda\beta^2)(\beta_g^2/\lambda)}
     =\beta\beta_g.
 \end{aligned}
\end{equation*}
\end{proof}

This concerns the \emph{full ellipsoid} and one linear block. Intersecting it
with a bridge class changes the support function and can reduce the width.
A null space forced by the stage-1 instrument alphabet need not persist in
later empirical blocks with many history cells. Population and empirical null
spaces must be analyzed separately; a dimension bound fixes neither the
orientation of an empirical null nor the leakage of a changing gradient.

\begin{proposition}[Pairwise uncertainty]
With the other blocks frozen, write each policy's affine value as
$V_i(b)=c_i+g_i^\top b$ and $V_j(b)=c_j+g_j^\top b$.
On the same region,
\begin{equation}
 \begin{aligned}
 \inf_{b\in C_\xi}\bigl(V_i(b)-V_j(b)\bigr)
 &=c_i-c_j+(g_i-g_j)^\top\hat b\\
 &\quad-\sqrt\xi\|g_i-g_j\|_{H^{-1}}.
 \end{aligned}
 \label{eq:pairwise}
\end{equation}
Thus positivity of the right-hand side certifies this blockwise ordering.
\end{proposition}
\begin{proof}
Apply the ellipsoid support formula to $g_i-g_j$ and retain the constant
$c_i-c_j$. It vanishes only when those offsets are equal.
\end{proof}

Large marginal widths do not establish ranking impossibility. For example,
let $b=(2,u)$, $|u|\le10$, $g_i=(1,1)$ and $g_j=(0,1)$.
Both marginal intervals have half-width 10, exceeding the value difference 2,
yet $V_i-V_j=2$ throughout the region. The common unidentified component
cancels. An impossibility of ranking would require observationally
indistinguishable admissible models with different policy orderings, which we
have not constructed. The coupled multiblock comparison is generally
multilinear and is not certified by Eq.~\eqref{eq:pairwise} alone.

\begin{proposition}[Candidate addition and relative score error]
Fix true values $V$ and finite scores $s$ on nonempty finite candidate pools
$\mathcal P\subseteq\mathcal Q$, keeping each existing candidate's score
unchanged. Let $i\in\arg\max_{\mathcal P}s$ and
$j\in\arg\max_{\mathcal Q}s$, and define
$e(\pi)=s(\pi)-V(\pi)$ and
$R_{\mathcal Q}(\pi)=\max_{\rho\in\mathcal Q}V(\rho)-V(\pi)$.
Then
\begin{equation}
\begin{aligned}
R_{\mathcal Q}(i)-R_{\mathcal Q}(j)
 &=s(j)-s(i)+e(i)-e(j)\\
 &\ge-\bigl(\max_{\mathcal Q}e-\min_{\mathcal Q}e\bigr).
\end{aligned}\label{eq:addition}
\end{equation}
If $|e|\le\epsilon$, candidate addition can worsen regret by at most
$2\epsilon$. If all scores are valid lower bounds, the possible harm is at
most $V(i)-s(i)$, but need not be zero.
\end{proposition}
\begin{proof}
The common regret reference cancels, and substituting $V=s-e$ gives the
identity. Nested maximization implies $s(j)\ge s(i)$, giving the inequality.
For lower scores, write $a=V-s\ge0$: then
$V(i)-V(j)=s(i)-s(j)+a(i)-a(j)\le a(i)$.
\end{proof}
This is elementary finite-pool score-comparison algebra, not a new
identification theorem. Common score bias cancels; relative errors matter.
Even valid lower scores allow harm: an old candidate with
$(V,s)=(.8,.1)$ loses to a new one with $(V,s)=(.3,.2)$.
Accordingly, a clone is a diagnostic candidate, not a guaranteed repair.

\paragraph{Moment-consistent relaxation.}
For a consistent exact moment equation $Lb=y$ (so $y\in\operatorname{range}L$),
write $b_0=L^\dagger y$ and assume
$\|b\|\le M$, with $M\ge\|b_0\|$. The half-width of $g^\top b$ over this
relaxed set is
\begin{equation}
 \|P_{\ker L}g\|\sqrt{M^2-\|b_0\|^2}.
\end{equation}
If $b_\star$ solves those same exact moments, choosing $M=\|b_\star\|$
gives $\|P_{\ker L}g\|\|P_{\ker L}b_\star\|$. This choice of $M$ is
necessary for that equality when $\|P_{\ker L}g\|>0$; when it is zero,
both expressions vanish for every admissible $M$. The half-width need not
equal the identified interval over valid POMDP primitives. In particular, the
true bridge norm and latent model are not available to a practitioner before
fitting.

\paragraph{Rate interpretation.}
With $\lambda=\lambda_0 N_2^{-\kappa}$ and
$\xi=cN_2^{\tau-1}/m$, the null contribution to the width is exactly
\begin{equation}
 W_{\mathrm{null}}=\sqrt{c/(m\lambda_0)}
 N_2^{(\tau+\kappa-1)/2}\|\Pn g\|.
\end{equation}
Growth follows only if the null-gradient norm does not decay fast enough;
an exponent alone does not determine regret. Bounded classes also prevent
unbounded value excursions. We use $(\tau,\kappa)=(0,1.5)$ as a diagnostic
schedule; it is not an admissible instantiation of the published theoretical
schedule and carries no theoretical coverage claim.

\section{Experimental protocol}
\paragraph{Environment.}
The primary setting has $|S|=4$, $|O|=6$, $|O_0|=2$, $|A|=2$, horizon $T=3$,
and environment seed 0. Emissions, negative controls and transitions are
random stochastic matrices with a diagonal boost. Rewards are Bernoulli;
action $s\bmod2$ has mean .85. The logger is
$\pi_b(a\mid s)=(1-q)/2+q\mathbf1\{a=s\bmod2\}$ with $q=.9$.
Emission and control matrices have maximal \emph{possible} rank; the control
cannot have full row rank in this incomplete configuration.

\paragraph{Estimation and calibration.}
Each simulation generates $N$ trajectories and splits them 70/30 into the two
estimation stages. Delta-kernel conditional mean embeddings and a quadratic
stage-2 ridge solve produce reward and dynamic bridges. For schedule C,
$\lambda_2=N_2^{-1.5}$ and the implementation uses
$\xi=c/(N_2\hat\sigma_{\mathrm{signal}})$, with an eigengap-based signal
estimate. At $N=2000$, seed 0, $c$ is chosen to give the first reward block
a penalty .12 for the fixed \texttt{greedy\_lo} candidate, then held fixed.
This is heuristic calibration, not a confidence-level guarantee. The reference
policy and seed used for calibration are disclosed because they affect results.

\paragraph{Policies and optimization.}
The six stationary observation policies are \texttt{always\_0},
\texttt{always\_1}, \texttt{uniform}, \texttt{greedy\_lo},
\texttt{greedy\_hi}, and \texttt{soft}. With
$\ell_o$ equally spaced from .9 to .1, their action-0 probabilities are
$1,\allowbreak0,\allowbreak.5,\allowbreak\ell_o,\allowbreak1-\ell_o$,
and $.5+.5(\ell_o-.5)$, respectively.
The plug-in and pessimistic selectors share fitted bridges and the empirical
initial-observation distribution. Coupled pessimism uses at most 30 coordinate
sweeps, tolerance $10^{-11}$, and three random restarts plus a centre start.
Exact dynamic programming evaluates each selected observation policy.
\emph{Selection regret} is relative to the best candidate in the evaluated set,
not the optimal unrestricted or latent-state policy.

Writing $\mathcal P_6$ for the fixed candidates and
$\mathcal P_{7,r}=\mathcal P_6\cup\{\hat\pi_{\mathrm{BC},r}\}$ for seed $r$,
the two selectors maximize their plug-in or heuristic pessimistic scores
over the specified pool. In the logged-data control every arm, including
selection restricted to $\mathcal P_6$, is evaluated on the same scale:
\[
 R_r(\hat\pi)=\max_{\pi\in\mathcal P_{7,r}}V_\star(\pi)
                 -V_\star(\hat\pi).
\]
Thus the candidate-addition contrast compares selected true values without
changing the regret reference within a seed. It measures an intervention on
the available choices, not an improvement to the bridge estimator.

Each block solve is exact for the unconstrained ellipsoid. Coordinate descent
need not find the joint global minimum; agreement between restarts is a
diagnostic, not a certificate. Nor does a score below ground truth certify
region coverage. Projection remedies can exclude true bridges and are not
certified alternatives.

\section{Evidence and the candidate-set intervention}
\begin{table}[t]
\centering\footnotesize
\caption{Schedule C: mean regret $\pm$ 95\% Student-$t$ Monte Carlo
half-interval over 20 trajectory seeds in one fixed environment. Regret is
relative to the best of the six fixed candidates; the last column uses
within-seed differences. Intervals are descriptive, not coverage guarantees.}
\begin{tabular}{rrrr}\toprule
$N$ & pessimistic & plug-in & paired difference\\\midrule
2,000 & $.09517\pm.02080$ & $.08937\pm.02072$ & $.00581\pm.02214$\\
8,000 & $.12185\pm.02238$ & $.02256\pm.01476$ & $.09929\pm.02501$\\
32,000 & $.17142\pm.01888$ & $.00000\pm.00000$ & $.17142\pm.01888$\\
128,000 & $.18044\pm.00000$ & $.00000\pm.00000$ & $.18044\pm.00000$\\\bottomrule
\end{tabular}\label{tab:fixed}
\end{table}
Table~\ref{tab:fixed} isolates the selection layer on shared estimates.
At the largest sample size, every one of the 20 stored runs selects
\texttt{uniform} pessimistically and \texttt{always\_1} by plug-in.
Zero dispersion across these runs does not prove correctness or establish
generalization across environments. The experiment varies trajectory seeds
while holding the environment fixed.
The smallest-sample paired difference is not separated from zero by this
interval approximation. Table~\ref{tab:fixed} recomputes these summaries
from the stored selections, using sample standard deviations.

\paragraph{Oracle candidate addition.}
The existing seventh candidate is computed from known simulator primitives:
\begin{equation}
 \pi_{\mathrm{oracle\text{-}BC}}(a\mid o)
 =\sum_s\frac{p_1(s)E(s,o)}{\sum_{s'}p_1(s')E(s',o)}\pi_b(a\mid s).
\end{equation}
It is the population stage-1 observation-marginal logger, reused as a
stationary policy. It is not a finite-data baseline learned from the logs.
It slightly exceeds the best original candidate's true value, by .000926.
With it included, stored pessimistic mean regret is zero at the first three
sample sizes and $\OracleBCAtLarge$ at $N=128000$, with a reported normal
interval half-width .0318. The modal selection is the oracle clone at every
sample size. These intervals used population-standard-deviation normalization
and are descriptive Monte Carlo summaries, not calibrated confidence statements
about a new environment. Raw selections were not retained for this legacy run,
so a paired uncertainty analysis cannot be recovered from its means alone.
This is a legacy diagnostic: its generator is supplied, but it was not rerun
as part of the fresh 80-fit logged-data release verification.

The fixed-set result is not universal. Adding this candidate reduces the
largest-sample mean about fivefold and changes the reference optimum.
It does not show that similarity to a logger guarantees high value. This
reward design makes the action preferred by the logger high-reward.

\paragraph{Logged-data control.}
The new reproduction driver estimates a stationary clone from only stage-1
training-split pairs $(O_1,A_1)$ using .5 pseudocounts per action. It requires no
latent states, emissions or true behaviour probabilities. It records every
seed's selected policy, exact value, score and restart gap, reports a common
augmented-set reference across selectors, and uses paired differences with
Student-$t$ Monte Carlo intervals. Table~\ref{tab:learned} gives the completed
20-seed reproduction. At $N=128000$, adding the learned candidate changes full
selection regret from $.1814$ to $\LearnedPessAtLarge$ on the \emph{same}
augmented reference. The paired difference between full and plug-in selection
is $.03630\pm.03486$. The pessimistic selector chooses \texttt{uniform} in
4 of 20 runs and the learned clone in the other 16. The maximum restart
gap over all 560 candidate evaluations is $\MaximumRestartGap$ (rounded);
small gaps still do not
certify global optimality.
An additional optimizer-budget check recomputes all seven candidate scores
for the 20 original largest-$N$ fits using 30 restarts and 150 sweeps.
It changes no selected policy; the largest absolute score change is
$1.136\times10^{-9}$. This reduces concern about this particular budget,
without certifying the global minimum or confidence-region coverage.
The within-seed reduction from adding the candidate is
$\CandidateGainAtLarge\pm\CandidateGainHalfInterval$ under the same
Student-$t$ summary. This directly quantifies the candidate-set intervention;
its uncertainty is conditional on this fixed environment and calibration.

\begin{table}[t]
\centering\small
\caption{New logged-data reproduction: mean candidate-relative regret and
95\% Student-$t$ Monte Carlo half-intervals across 20 trajectory seeds in one
fixed environment. Every arm uses the same per-seed augmented optimum.
Negative interval endpoints, where present, are approximation artifacts.}
\begin{tabular}{rrrr}\toprule
$N$ & full, learned candidate & plug-in, learned candidate & clone alone\\\midrule
2,000 & $.00297\pm.00179$ & $.00297\pm.00179$ & $.00297\pm.00179$\\
8,000 & $.00140\pm.00084$ & $.00130\pm.00085$ & $.00140\pm.00084$\\
32,000 & $.00057\pm.00047$ & $.00057\pm.00047$ & $.00057\pm.00047$\\
128,000 & $.03632\pm.03486$ & $.00002\pm.00005$ & $.00002\pm.00005$\\\bottomrule
\end{tabular}\label{tab:learned}
\end{table}

The practical control reproduces the candidate-set sensitivity of the oracle
diagnostic. It also shows that cloning alone is competitive in this
construction. We do not infer that the modest residual difference is stable
across other latent environments or candidate-generation schemes. The intervals
describe trajectory randomness conditional on environment seed 0; they have no
simultaneous coverage guarantee across all reported sample sizes.

\subsection{Transferred calibration on new environments}
We next freeze the original multiplier, without fitting or selecting it on any
new environment. Environment seeds $1,\ldots,5$ use the same generator,
dimensions, $q=.9$ and $T=3$; trajectory seeds $20,\ldots,29$ are disjoint
from the original seeds. We run $N\in\{8000,128000\}$, yielding 100 new
fits and 700 scored candidates. Each within-seed contrast retains identical
fits and the augmented regret reference. Exact backward evaluation is checked
against independent forward latent-state propagation.

\begin{table}[t]
\centering\small
\caption{Transferred calibration: mean regret $\pm$ descriptive 95\%
Student-$t$ half-interval across five environment means (10 trajectory
seeds per environment). These are five environment units, not 50 independent
environments. Candidate gain is the paired six-minus-augmented regret.}
\begin{tabular}{rrr}\toprule
Quantity & $N=8000$ & $N=128000$\\\midrule
Pessimistic, six & $.21087\pm.10790$ & $.26069\pm.07892$\\
Pessimistic, augmented & $.04162\pm.08727$ & $.03961\pm.09666$\\
Plug-in, augmented & $.03245\pm.09008$ & $.03265\pm.09066$\\
Clone alone & $.03245\pm.09008$ & $.03265\pm.09066$\\
Candidate gain & $.16925\pm.16136$ & $.22109\pm.12082$\\\bottomrule
\end{tabular}\label{tab:transfer}
\end{table}

At $N=128000$, the candidate improves mean pessimistic selection in all five
new environments. At $N=8000$ it helps in four and harms in one: in
environment 4 mean regret increases from $.15653$ to $.16223$.
Thus the added candidate is not a monotone improvement, consistent with
Eq.~\eqref{eq:addition}. At the larger $N$, environment 4 has clone and
plug-in regret $.16327$; nearly perfect plug-in performance in environment 0
is not replicated across the generator. The across-environment paired
pessimistic-minus-plug-in difference is $.00696\pm.01205$ and is not
separated from zero by this approximation. We therefore emphasize candidate
sensitivity rather than a universal superiority ranking of selectors.

\subsection{Reward-reversal stress control}
The original generator makes the logger's preferred action high-reward,
which can favor cloning. Before inspecting this control's outcomes, we fix
the following intervention: in each new environment $1,\ldots,5$, exchange
the two action columns of the reward-mean matrix, keeping the emissions,
transitions, logger, horizon and original calibration unchanged. This is a
different reward design, not a change to the logged action mechanism.
We use fresh trajectory seeds $30,\ldots,34$ at $N=128000$ (25 new fits).
Each arm still uses common fits and the same augmented regret reference.

\begin{table}[t]
\centering\small
\caption{Reward reversal, $N=128000$: descriptive summaries over five
environment means, each with five trajectory seeds. Candidate gain is
six-minus-augmented regret; negative values indicate harm.}
\begin{tabular}{rr}\toprule
Quantity & Mean $\pm$ Student-$t$ half-interval\\\midrule
Pessimistic, six & $.13718\pm.19219$\\
Pessimistic, augmented & $.26915\pm.24003$\\
Plug-in, augmented & $.16645\pm.26242$\\
Clone alone & $.43142\pm.15260$\\
Candidate gain & $-.13197\pm.13595$\\\bottomrule
\end{tabular}\label{tab:reversal}
\end{table}

Adding the clone increases mean pessimistic regret in four environments and
leaves it unchanged in one. The mean increase is $.13197$; its wide interval
includes zero, so we do not claim a population-level sign. Clone-only regret
is $.43142$, showing why cloning is a poor default in this stress design.
Together the two interventions support candidate-set sensitivity in both
directions, not a claim that adding a logged-data clone reliably repairs
pessimistic selection. The reversed-reward case is a deliberately constructed
stress control rather than a naturally occurring clinical or robotic benchmark.

\subsection{A real-outcome stroke-trial boundary case}
To examine medical relevance without inventing patient counterfactuals, we
add a single-decision ($T=1$) illustration using the public corrected version
2 of the International Stroke Trial (IST)~\cite{istdata,isttrial}.
Artificial training selection and held-out IST evaluation have precedent in
Kallus and Zhou~\cite{kalluszhou}; our treatment, endpoint and proxy construction differ.
Of 19,435 patients, 19,433 have ascertained 14-day vital status
(\texttt{SET14D}=1). We use randomized aspirin \emph{assignment}
(\texttt{RXASP}), not received treatment or compliance; the binary reward is
14-day survival ($1-\mathrm{ID14}$). The estimand averages over the factorial
heparin coassignment. No post-randomization diagnosis or adherence exclusion
is made. The two unknown outcomes are excluded and reported.

Before fitting, one fixed patient-disjoint split (seed 20261111) allocates
9,716 patients to training and 9,717 to evaluation. The observation is the
six-category cross of sex and baseline consciousness. We hide the baseline
indicator $U=\mathbf1\{\mathrm{age}\ge75\}$ from the policy learner and
generate $O_0=U\mathbin{\mathrm{XOR}}\mathrm{Bernoulli}(.2)$ with independent
noise. This is an \emph{artificial proxy}, not an established clinical negative
control. In the selection-biased training condition, a patient is retained
with probability .8 if assigned action equals $U$, and .2 otherwise; outcome
is not used in retention. An unfiltered-training control is also fitted.
We use 20 training perturbations (seeds 100--119), yielding 40 fits; these
reuse patients and are not independent clinical replications. Between 4,776
and 4,910 patients remain in each selection-biased fit.

The same six stylized computational policies and a training-split clone are
scored with $T=1$ bridges and the original fixed multiplier. The initial
observation marginal comes from \emph{unfiltered training} covariates, preserving
the target population rather than the retained population. The one-block
ellipsoid minimum is exact here; no coupled optimizer is needed. Clinical
bridge validity and completeness are not established. The policies are not
proposed clinical treatment rules.

We average each arm's 20 selected policies and evaluate that fixed mixture
on untouched trial patients. For a policy $\pi$, the working reference is
\[
 \widehat V_g(\pi)=\frac1{n_{\rm test}}\sum_{i\in\rm test}
 \frac{Y_i\pi(A_i\mid O_i)}{g(A_i\mid X_i)}.
\]
For a comparison, paired patient contributions replace $\pi$ by
$\pi_{\rm augmented}-\pi_{\rm six}$. We report their mean and normal
half-interval $1.96s/\sqrt{9717}$, conditional on fitted policies and weights.
IST used minimisation in randomization~\cite{isttrial}; equal marginal
allocation does not establish a known individual propensity of .5.
Thus $g(1\mid X)=.5$ is explicitly a working equal-allocation reference.
A sensitivity analysis estimates propensity from unfiltered training only,
using 90 baseline strata (age band, consciousness, stroke subtype and sex),
Beta$(1,1)$ smoothing and clipping to $[.1,.9]$. Neither reference supplies
exact counterfactual truth or a design-certified causal regret estimate.

\begin{table}[t]\centering\small
\caption{IST: paired held-out value change from candidate addition; positive
means a larger estimated 14-day survival assignment value. Half-intervals
condition on the fitted 20-policy mixture, not 20 independent patient samples.}
\begin{tabular}{llr}\toprule
Training & Evaluation weight & Value change $\pm$ half-interval\\\midrule
Unfiltered & Working .5 & $.000052\pm.000106$\\
Unfiltered & Baseline sensitivity & $-.000045\pm.000111$\\
Selection-biased & Working .5 & $-.000252\pm.001351$\\
Selection-biased & Baseline sensitivity & $-.000283\pm.001360$\\\bottomrule
\end{tabular}\label{tab:ist}
\end{table}

All intervals in Table~\ref{tab:ist} include zero. The tiny unfiltered contrast
changes sign under propensity sensitivity; no medical value improvement is
resolved. The maximum inverse observed propensity is 6 in the sensitivity
analysis. At working weight .5, the two excluded outcomes can change a
full-cohort paired mean by up to $2(2)/19435\simeq.000206$, also material
relative to these contrasts. This is real patient-outcome evidence for a
controlled \emph{semi-synthetic} selection-bias test, not validation of sequential
clinical RL, naturally available proxies, or treatment effectiveness.

\section{Limitations and practical implications}
The sequential evidence comes from one synthetic generator. It does not
establish general superiority or failure. Hand-chosen policies, seed-0 calibration,
an incomplete instrument, omission of the bounded bridge class, and an
uncertified joint optimizer all limit interpretation. No claim is made about
global optimal policy regret, deployment safety, or a pre-fit test usable
without access to latent primitives.

History length alone does not guarantee identification; transition and proxy
rank conditions still matter. Neither marginal-width diagnostics nor small
restart gaps establish policy-ordering or coverage guarantees. The exact
feasible class, calibration and joint optimization must be assessed together.

For an interpretable comparison, the useful checks are operational: disclose the exact
feasible class and calibration; compare policies on identical fits; include a
clone learned from logged data; retain per-seed selections; and report when
the signal-width denominator vanishes. These checks narrow a negative result
into an interpretable case study.
The IST illustration broadens the data source but is only $T=1$, has constructed
selection and proxy channels, and gives estimated trial-reference values with
working weighting assumptions. Its normal intervals omit training and adaptive
randomization uncertainty. It cannot resolve the sequential study's
identification, clinical transportability or confidence-calibration limits.

\section{Conclusion}
A calibrated, class-free proximal pessimism implementation selects worse
candidates as data increases in the original incomplete tabular design.
Candidate addition improves selection in five new environments at the larger
sample size, but harms it in four of five reward-reversal environments.
Thus a logged-data clone is a useful sensitivity diagnostic, not a universal
remedy. The blockwise width floor is valid under its stated geometry;
policy ordering requires a pairwise or joint analysis. The defensible
contribution is an empirical sensitivity study with explicit implementation
limits, rather than a general impossibility or criticism of the anchor theorem.
The real-outcome IST boundary case resolves no survival-value improvement;
it supports transparent evaluation of selection sensitivity rather than
medical efficacy or validation of sequential clinical deployment.

\paragraph{Use of AI tools.}
Anthropic Claude (Fable and Opus) assisted with research code development.
OpenAI Codex assisted with code development, experiment execution, source
checking, and manuscript drafting and revision. Reported results are linked
to retained outputs and reproducibility checks. The authors take
responsibility for the manuscript's content.

\appendix
\section{Reproduction details}
\label{app:reproduction}
\paragraph{Random environment construction.}
All primitives are generated once with NumPy's default random generator at
seed 0. Emission rows are drawn from a symmetric Dirichlet with parameter .7,
then 1.5 is added to entry $s\bmod |O|$ and each row is renormalized. Negative
control rows use the same construction on $|O_0|$ categories and independent
draws. For each $(a,s)$, the transition row is drawn from a symmetric Dirichlet
with parameter .8, then 1.2 is added to entry $(s+a)\bmod |S|$ and the row is
renormalized. Reward means are drawn uniformly on $[.15,.85]$, with entry
$(s,s\bmod2)$ overwritten by .85. The initial latent distribution is drawn
from a symmetric Dirichlet with parameter 2. The generation order is emissions,
controls, transitions, rewards, then initial distribution. There is no discount:
values are expected sums of the three rewards.

\paragraph{Bridge fit.}
At stage $t$, write $W_t=(A_t,O_t)$,
$X_t=(A_t,H_{t-1},O_0)$, $Y_t=(R_t,O_t)$ and
$Z_t=(O_{t+1},O_t)$. The history $H_{t-1}$ contains preceding observations and
actions. Reward and dynamic bridges target the conditional probabilities of
$Y_t$ and $Z_t$, respectively. For training counts $n_1$, the first-stage
embedding is
\[
 \hat\mu(w\mid x)=\frac{n_1(w,x)}{n_1(x)+N_1\lambda_1},
 \qquad \lambda_1=1/N_1.
\]
Targets use conditional frequency estimates, with marginal fallback when
$n_1(x)=0$. Such unseen cells have zero embedding and do not contribute a
design vector. For each second-stage trajectory, an independent index $j_n$
is drawn uniformly from the reward or dynamic target alphabet. Set
$\psi_n=\hat\mu(\cdot\mid x_n)\otimes e_{j_n}$ and
$u_n=\hat p(j_n\mid x_n)$. The primal ridge fit solves
\[
 A=\frac1{N_2}\sum_n\psi_n\psi_n^\top,\qquad
 (A+\lambda_2 I)\hat b=\frac1{N_2}\sum_n\psi_n u_n.
\]
The three reward blocks have 144 coefficients each; the two dynamic blocks
have 432 each. The split permutation and index draws use the estimator's
simulation seed. The empirical distribution of $O_1$ used in value evaluation
is computed on all generated trajectories; the learned clone uses only the
first-stage training split.

\paragraph{Width and optimization.}
For each block and action, form the occupancy-weighted embedding covariance
$M_aM_a^\top/N_2$. Sort its eigenvalues $d_1\ge\cdots\ge d_6$ and choose
\[
 k=\arg\max_{1\le i<6}\frac{\max(d_i,f)}{\max(d_{i+1},f)},
 \qquad f=10^{-9}\max(d_1,0).
\]
If $f=0$, use $k=1$.
The signal estimate is the minimum over actions of $d_k$.
For the reported runs the multiplier is $c=4.151795352832782\times10^{-6}$.
It is held fixed across sizes and seeds; zero or negative signal estimates
make this width undefined in the logged-data and new sensitivity drivers.
The historical fixed-set and oracle drivers instead floor the signal
denominator at $10^{-12}$; their degenerate-signal handling must not be
interpreted as valid confidence calibration.
Reward blocks are updated together at fixed dynamic blocks; dynamic blocks
are then updated sequentially with refreshed gradients. Random restarts use
generator seed 0 for each candidate evaluation. Ties in policy scores are
broken by the insertion order
\texttt{greedy\_lo}, \texttt{greedy\_hi}, \texttt{always\_0},
\texttt{always\_1}, \texttt{uniform}, \texttt{soft}, \texttt{learned\_bc}.

\paragraph{Clone and uncertainty summaries.}
With first-stage training counts $n_1(o,a)$, the learned policy is
\[
 \hat\pi_{\mathrm{BC}}(a\mid o)
 =\frac{n_1(o,a)+.5}{\sum_{a'}n_1(o,a')+1}.
\]
It is reused at every stage. For each seed, true candidate values are computed
by backward dynamic programming from simulator primitives, used only for
evaluation. If $r_i$ is candidate-relative regret, the table reports
$\bar r\pm t_{.975,19}s_r/\sqrt{20}$ with sample standard deviation $s_r$.
Paired comparisons apply this expression to the within-seed regret differences.

\paragraph{Computational record.}
The new experiment uses Python 3.12.14, NumPy 2.5.3 and SciPy 1.18.1.
The reproduction files are:
\begin{flushleft}\small
Driver: \path{Extension/poc/run_learned_bc.py}\\
Results: \path{Extension/experiments/results_learned_bc.json}
\end{flushleft}
The grid is $N\in\{2000,8000,32000,128000\}$ and seeds $0,\ldots,19$.
Stored records contain policy probabilities, scores, selections, exact values,
regrets and restart gaps. The separate numeric check
\texttt{verify\_workshop\_numbers.py} recomputes regrets and uncertainty
summaries from these records and compares them to the manuscript.
These checks verify internal consistency; they do not substitute for independent
replication or establish coverage of the heuristic confidence regions.
A fresh extraction of the reproduction archive regenerated all 80 fitted
cases and matched the supplied raw scores, selections, regrets and statistical
summaries at relative/absolute tolerance $10^{-8}$.
The transfer and reward-stress drivers are:
\begin{flushleft}\small
\path{Extension/poc/run_external_validity.py}\\
\path{Extension/poc/run_reward_reversal.py}
\end{flushleft}
Both retain every score,
selection, regret and environment-level summary. The fixed designs use
100 and 25 new fits, respectively; environment-level intervals use
$t_{.975,4}$ on the five environment means, not the trajectory count.
Additional drivers reproduce the optimizer diagnostic and the IST experiment:
\begin{flushleft}\small
\path{Extension/poc/run_optimizer_budget.py}\\
\path{Extension/poc/run_ist_policy_selection.py}
\end{flushleft}
The IST downloader verifies the corrected source checksum. Patient-level data are not
included in the companion; download them from the cited original repository.
Contains information from the International Stroke Trial database (version 2),
made available under the Open Data Commons Attribution License (ODC-By 1.0),
\url{https://opendatacommons.org/licenses/by/1-0/}. This data licence is separate
from the MIT code licence. The stored experiment record contains aggregate
counts, scores and six-by-two policies, not patient records.
\paragraph{Artifact availability.}
The named release includes the reproduction code, raw JSON records,
reference-version constraints and MIT code license as ancillary material
prepared for distribution with this preprint. Relative file paths are
preserved. Download and extract the ancillary files together, then run the
README commands.
The historical oracle diagnostic is distinguished from the fully rerun
logged-data experiment. The project repository is
\href{https://github.com/soheilsayahvarg/model-based-rl-confounded-pomdps}{available on GitHub}.
\end{document}